\documentclass[letterpaper, 10 pt, conference]{ieeeconf}  

\IEEEoverridecommandlockouts                              

\usepackage{cite}
\usepackage{amsmath,amssymb,amsfonts}
\usepackage{algorithmic}
\usepackage{graphicx}
\usepackage{textcomp}
\usepackage{xcolor}

\usepackage{xspace}
\usepackage{subfig}
\usepackage[AMmath, AMcolor]{AMlatex}
\usepackage{units}
\usepackage{balance}
\usepackage{gensymb}
\usepackage{tikz}
\usepackage{url}
\usepackage{booktabs}

\makeatletter
\let\NAT@parse\undefined
\makeatother
\usepackage{hyperref}

\usepackage{cleveref}

\crefname{equation}{}{}
\crefname{figure}{Fig.}{Figs.}
\crefname{table}{Table}{Tables}
\crefname{algorithm}{Algorithm}{Algorithms}
\crefname{appendix}{App.}{App.}
\crefname{section}{Section}{Sections}
\crefname{subsection}{Subsection}{Subsections}

\newcommand{\lola}{\textsc{Lola}\xspace}

\usepackage{import}

\usepackage{pgfplots}
\usepackage{pgfgantt}
\pgfplotsset{compat=default}

\usetikzlibrary{spy}
\usetikzlibrary{shapes,arrows,fit}
\usetikzlibrary{calc}
\usetikzlibrary{patterns}
\usetikzlibrary{external}
\pgfplotsset{compat=1.16}

\title{\LARGE \bf
Learning the Noise of Failure: Intelligent System Tests for Robots
}

\author{Felix Sygulla and Daniel Rixen$^1$
\thanks{$^{1}$Technical University of Munich, Chair of Applied Mechanics, Boltzmannstr. 15, 85748 Garching, Germany
        {\tt\small felix.sygulla@tum.de}}%
}
\begin{document}
\maketitle
\thispagestyle{empty}
\pagestyle{empty}

\begin{abstract}
Roboticists usually test new control software in simulation environments before evaluating its functionality on real-world robots. Simulations reduce the risk of damaging the hardware and can significantly increase the development process's efficiency in the form of automated system tests.

However, many flaws in the software remain undetected in simulation data, revealing their harmful effects on the system only in time-consuming experiments. In reality, such irregularities are often easily recognized solely by the robot's airborne noise during operation. We propose a simulated noise estimate for the detection of failures in automated system tests of robots. The classification of flaws uses classical machine learning ---  a support vector machine --- to identify different failure classes from the scalar noise estimate.

The methodology is evaluated on simulation data from the humanoid robot \lola. The approach yields high failure detection accuracy with a low false-positive rate, enabling its use for stricter automated system tests. Results indicate that a single trained model may work for different robots. The proposed technique is provided to the community in the form of the open-source tool \emph{NoisyTest}, making it easy to test data from any robot. In a broader scope, the technique may empower real-world automated system tests without human evaluation of success or failure.
\end{abstract}

\section{Introduction}

Experimental testing of new control algorithms for robots is often time-consuming and sometimes a safety issue. However, it is common to reduce the time spent on experiments and run simulated system tests of the robot with virtual hardware and environment first. The robot can be tested safely by simulating its movement using full multi-body dynamics, environment model, and actuator/sensor characteristics. Failure evaluation can either be done by manual inspection of the simulation results or as part of automated system tests in combination with distinct conditions for failure or success. Based on the granularity of the used simulation models, the results can be quite close to real-world observations --- although there will certainly be unmodeled effects, which require final experimental tests.

\begin{figure}
	\centering
	\resizebox{6cm}{!}{\includegraphics{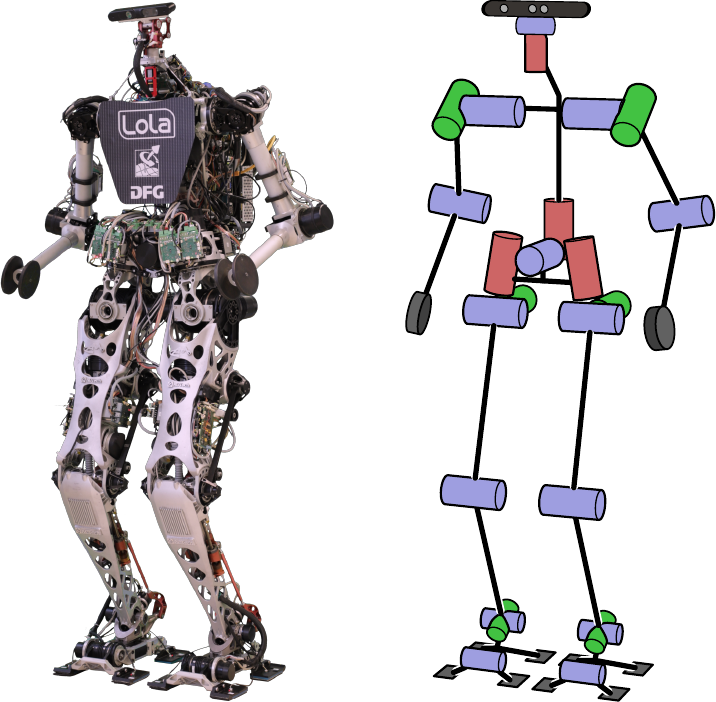}}
	\caption{The humanoid \lola and its kinematic structure.}
	\label{fig:lola}
\end{figure}

From our experience with testing algorithms on the biped robot platform \lola (\cref{fig:lola}), we found that simulation data often contains clues on flaws in the robot's behavior. Still, these remain undetected in an automated or manual inspection. In all these cases --- one could term them soft failures --- the robot reached its high-level goal, i.e., it did not fall or incline to an unexpected degree.
In the corresponding experiments, however, one could immediately hear something is wrong -- just from the machine's noise. The reason was, for example, an oscillating feedback loop, or hard touchdown of the feet. We also found that clues for these failures are hidden in the simulation results. So the question we want to answer in this paper is: Can we use that information for system testing?

We propose a method to assess simulated system tests based on an estimated noise of the simulated robot. The noise signal is analyzed using classical machine learning approaches to distinguish between normal and failure-related noise.
Typical fault detection and fault diagnosis approaches for robots focus on detecting failures during their use in a production environment --- caused by faulty sensors or actuators, \cite{Khalastchi2018}. Common techniques for manipulators use the residual of a model-based observer on the system's dynamics to detect and isolate actuator failures, \cite{Vemuri1998, Dixon2000, DeLuca2005}. To the authors' knowledge, there is no work on the use of these techniques for the assessment of system tests in the development stage. Vibration signals are, however, commonly analyzed with machine learning approaches for fault diagnosis of rotating machinery, \cite{Liu2018}. Typically, measured acceleration signals are transformed into the frequency domain, and the levels of individual subbands are then fed into neural networks or support vector machines, \cite{Martinez2011, Fernandez2013}. Alternative approaches use wavelet transform coefficients\cite{Guo2018} or a large set of features from classical signal analysis \cite{Santos2015} as input data. In contrast, the work presented here uses a simulated (acoustic) noise --- not a vibration (acceleration) signal --- as input. The preprocessing pipeline uses a short-time Fourier transformation of the estimated noise signal and compresses the resulting spectrum instead of using individual frequency subbands.

This paper's main contribution is a methodology to estimate the robot noise in simulation and detect soft failures based on features in this signal. The article is organized as follows: \cref{sec:noisytest:simulation} describes the used simulation environment and generated noise data set. In \cref{sec:noisytest:classification}, the preprocessing pipeline and classification approach based on support vector machines are explained. \cref{sec:noisytest:results} describes the conducted hyper-parameter search and presents the accuracy results on a validation data set. Furthermore, we discuss the generalizability of the approach. \cref{sec:noisytest:conclusion} concludes the article and proposes ideas for future work and use cases.

\section{Simulation Environment and Dataset}
\label{sec:noisytest:simulation}
The proposed failure detection method is applied to simulation data of the biped walking research platform \lola. The robot is actuated by 24 electro-mechanical drives; see \cref{fig:lola} for its kinematics structure. More details on the hardware design are described in \cite{Lohmeier2010, Sygulla2018}.

The custom simulation environment for \lola includes models of the full rigid multi-body dynamics, the contacts with the environment, and the drive modules --- consisting of a decentralized joint controller, electrical system, motor, and gearbox. For details we refer the reader to \cite{Buschmann2010, Favot2016}. The simulation software also includes instances of the \emph{Walking Pattern Generator (WPG)}\cite{Seiwald2019} and the \emph{Stabilization and Inverse Kinematics (SIK)}\cite{Sygulla2020} modules to control the humanoid in its virtual environment. For a typical simulation run, high-level commands are sent to the \emph{WPG} module to have the robot execute a certain movement (e.g. walk straight). In addition, obstacles (in this case unknown to the control software of the humanoid) can be defined for the virtual environment. Major outputs of the simulation software are the movement of all bodies of the robot --- that is the actual trajectories of all joints $\vq_j, \dot \vq_j, \ddot \vq_j \in \mathbb{R}^{24}$ --- and the trajectory of the floating robot base $\vq_{f}, \dot \vq_{f}, \ddot \vq_{f} \in \mathbb{R}^6$. The simulation output is already validated using several heuristics, to check, for example, that the robot did not fall during its motion.

\subsection{Noise Estimation}
In this paper, we use an estimated scalar noise signal calculated from raw simulation data to detect and classify possible soft failures of the machine. We assume the robot is built of rotational joints only. To derive our approximation of the noise radiated by the virtual robot, let's first consider a rigid body $i$ with surface $S_i$, which is attached to a parent body ${i-1}$ via the rotational degree of freedom $q_i$. The body $i$ rotates around the point $O_i$ of its parent. The radiated sound power of this body is then defined as \cite{Cremer2005} 
\begin{align}
\label{eq:p:correct}
P_i(t) = \rho c \sigma_i S_i v_{\bar S_i}^2(t),
\end{align}
with the density of the air $\rho$, the speed of sound $c$, the radiation efficiency $\sigma_i$, and the average velocity over all points $P$ of the body's surface
\begin{align}
v_{\bar S_i}(t) = \frac{1}{S_i} \int_{S_i} ||\vv_P|| \dd P.
\end{align}
The radiation efficiency describes losses caused from the body to air transition and the fact that the average surface velocity is not on all parts of the body perpendicular to its surface. Using rigid body kinematics, the average surface velocity can be rewritten as
\begin{align}
\label{eq:v:correct}
v_{\bar S_i}(t) = \frac{1}{S_i} \int_{S_i} \underbrace{||\vv_{O_i}||}_A &+ \underbrace{||\vomega_{i-1} \times \vr_{iP}||}_B \notag \\
& + \underbrace{||\vomega_{i-1,i} \times \vr_{iP}||}_C \dd P,
\end{align}
with the velocity of the pivot point $\vv_{O_i}$, the angular velocity of the parent body, $\vomega_{i-1}$, the relative angular velocity of the body, $\vomega_{i-1,i}$, which depends only on $\dot q_i$, and the position of a point $P$ on the surface of the body, $\vr_{iP}$. At this point, we scrutinize the exactness of the virtual noise signal for the sake of simplicity. The goal is to generate a signal, which does contain all information of the correctly calculated noise, but may deviate in scaling and amplitude. The terms $A$ and $B$ in \cref{eq:v:correct} do not depend on the degree of freedom $q_i$ of the considered body $i$, but on the movement of all parent bodies of $i$. We drop these terms and consider only the relative movement of every body in its surface velocity quantifier $\hat v_{\bar S_i}$. Furthermore, term $C$ can be simplified, because the rotation of a link is one-dimensional. This yields
\begin{align}
\label{eq:v:simple}
\hat v_{\bar S_i}(t) = \frac{1}{S_i} \int_{S_i} \dot q_i \; l_P \dd P = \dot q_i \; \underbrace{\frac{1}{S_i} \int_{S_i} l_P \dd P}_{G_i},
\end{align}
with the lever arm $l_P$ and a scalar $G_i$ characterizing the geometry of body $i$. By inserting \cref{eq:v:simple} into \cref{eq:p:correct} and setting all geometry-depending factors to 1 (we don't care about correct scaling), we retrieve the sound power quantifier
\begin{align}
\hat P_i(t) = \dot q_i^2 (t)
\end{align}
for body $i$. The total sound power quantifier for all $n$ joints can be written as
\begin{align}
\bar P(t) = ||\dot \vq||^2 (t),
\end{align}
with $\dot \vq \in \mathbb{R}^n$. This already gives an estimation of the radiated sound power caused from the actual movement of the robot's joints. The scaling between the noise coming from the individual joints is lost with this approximation. We are, however, interested in time-local features of the noise, not the relative weighing between certain noise sources. In addition, we add the sound power related to the desired movement of the robot to get
\begin{align}
	\label{eq:P:allbodies}
	\hat P(t) = ||\dot \vq||^2 (t) + ||\dot \vq^d||^2 (t),
\end{align}
with the desired joint velocities $\dot \vq^d \in \mathbb{R}^m$. This ensures that features only visible in the desired movement --- $\vq_j$ will in general not follow the desired $\vq^d_j$ exactly --- are also part of the noise estimate. By assuming radial and even distribution of this sound power, the observed pressure at a large distance $d$ to the bodies is \cite{Cremer2005}:
\begin{align}
p(t) = \sqrt{\hat P(t) \frac{\rho c}{4 \pi d^2}}.
\end{align}
By setting all scaling constants to $1$ and inserting \cref{eq:P:allbodies}, we obtain the final estimated noise pressure quantifier
\begin{align}
\hat p(t) = \sqrt{|| \dot \vq ||^2 + || \dot \vq^d ||^2}.
\end{align}
In the case of \lola, we use $\vq = [\vq_j, \vq_f]^T \in \mathbb{R}^n$, $\vq^d = \vq^d_j \in \mathbb{R}^m$ for the noise estimation with $n=\unit[30]{}$, $m=\unit[24]{}$.
\subsection{Considered Scenarios and Failure Classes}
\begin{table*}[tb]
	\centering
	\begin{tabular}{@{}p{0.7cm}p{2.2cm}p{6.7cm}p{5cm}p{1.4cm}@{}}\toprule
		Nr & Scenario\newline Description & Intentional Defect & Parameters & Contained Labels \\
		\midrule
		1 & Walking straight & - & - & \emph{OK} \\
		2 & Curve walking & - & - & \emph{OK} \\
		3-4 & Walking sideways, forward, backward, on the spot & - & Variation of step sequence and step duration & \emph{OK} \\
		5-9 & Walking over\newline an undetected\newline obstacle & Unexpected positive ground height change (obstacle) & Variation over different initial contact points with the obstacle (full foot, heel only, toe only, side only) resulting in different impact times in the simulation data & \emph{OK}, \emph{Impact}, \emph{HighAcc}\\
		10 & Walking straight & Decalibrated right ankle joint; the right foot is inclined on touchdown & \unit[0.05]{rad} calibration error & \emph{OK}, \emph{Impact}\\
		11-19 & Walking straight & Multiple sine-wave oscillations superposed to the vertical task-space position of the feet & Multi-sine with \unitfrac[1]{sine}{Hz}, amplitude: \unit[2]{mm}; several experiments with different frequency ranges:\unit[\{20-30, 30-40, 40-50, 50-60, 60-100, 100-150, 150-250, 250-350, 900-1000\}]{Hz} & \emph{Oscillations}\\
		20-21 & Walking straight & Unexpected positive ground change (obstacle) only on left foot; leads to repeated early and late contact & Height deviation: \unit[2]{cm}; variation of step height and step duration & \emph{Impact}\\
		22 & Walking down\newline a platform & Unexpected negative ground height change; the last step on the platform is a partial contact; causes the control scheme to execute dynamic foot motions with high accelerations. & \unit[6]{cm} platform height & \emph{HighAcc}, \emph{Oscillations}\\
		23-24 & Walking straight & Faulty force/torque sensor signals with 400N force or 30Nm torque signal jumps & - & \emph{HighAcc}, \emph{Oscillations}\\
		\bottomrule
	\end{tabular}
	\caption{The simulation scenarios and defects used for the generation of training and validation data}
	\label{tab:scenarios}
\end{table*}
We consider several test scenarios for the generation of training and validation data with the robot simulation. In all those scenarios, the robot does not fall; all these tests are considered as success by the implemented classical system tests. Still, the following undesired failure classes may occur in these simulations:
\begin{itemize}
	\item \emph{Hard impacts with the environment}: Faulty trajectories of the feet may lead to impacts on the ground - even for walking on level ground.
	\item \emph{Unusually high joint accelerations}: Badly parametrized control systems without limitation of the actuating variable may lead to velocity jumps or otherwise high accelerations.
	\item \emph{Oscillations}: Unstable control loops may lead to oscillations of the actuated bodies. The amplitude of these vibrations is usually small and does not cause the robot to fall --- still it is an undesired behavior.
\end{itemize}
Based on the failure classes, we define 4 possible labels for a block of time-domain noise signal data: $cls = \{$\emph{OK}, \emph{Impact}, \emph{HighAcc}, \emph{Oscillations}\}. The noise signal data is generated from simulation runs for the scenarios defined in \cref{tab:scenarios}.
The scenarios were designed to generate noise data for the failure classes in multiple ways, while still having a relation to real-world problems. The data is manually annotated in the time-domain, i.e. every time block of simulation data is linked to one label and has a different block length --- start and end time of a block is set manually based on the data. The total set of 83 time-data blocks corresponds to a total of \unit[45.7]{s} annotated noise signal training data, and \unit[27.6]{s} annotated noise signal validation data --- both at a sample rate of $f_s=\unit[10]{kHz}$. Note that in the scenarios 5-9, 20-21, and 22 the impact or high accelerations are actually expected and the simulations correspond to the best the control algorithms can do in these scenarios. Still, they are considered as failure training data, because their occurance in other situations may indicate a fault. We consider it up to a higher-level system test software to interpret \emph{NoisyTest}'s results depending on the context of the tested scenario.
%

\section{Classification Approach}
\label{sec:noisytest:classification}

This section describes the classification approach used to automatically learn the relations between the noise estimation signal and the annotated failure labels. In the following, several parameters for the preprocessing pipeline and classifier are introduced. Variables marked with a superscript $*$ are subject to a later hyper-parameter search, see \cref{sec:hyperparametersearch}; the values of all other parameters are directly motivated within this section.

\subsection{Data Preprocessing}

The preprocessing step transforms the estimated noise signal in time-domain to a meaningful, dense representation with preferably high relation to the failure classes. The structure of the preprocessing pipeline is motivated by techniques used in speech recognition systems. In the following, all steps of the pipeline are described.

\subsubsection{Framing}

The incoming labeled noise signal blocks all have a different duration / number of samples. By framing the data, all blocks are transformed to the same number of samples $n_f$. We expect failure patterns of the noise signal to fit in a time frame of $\Delta t_F = \unit[0.4]{s}$; consequently, we set the frame size to $n_f=\unit[4000]{samples}$. Given a data block with arbitrary size $n_c$, the following cases exist:
\par
(1) When $n_c \ge n_f$, we extract several overlapping frames of size $n_f$ using a stride length $s = \unit[1000]{samples}$ for a sliding window. This increases the total number of training frames and pertubates the time-domain data by time shifts. All extracted frames share the same label data of the original data block. Only windows which completely fit into the larger data block are extracted. The parameters $n_f$ and $s$ are set based on expert knowledge of possible data patterns and a coarse parameter search. They directly influence the size and composition of the data set and are considered fixed in the following.
\par
\begin{figure*}
	\centering
	\hspace{-1.5cm}
	\subfloat[]{
	\includegraphics{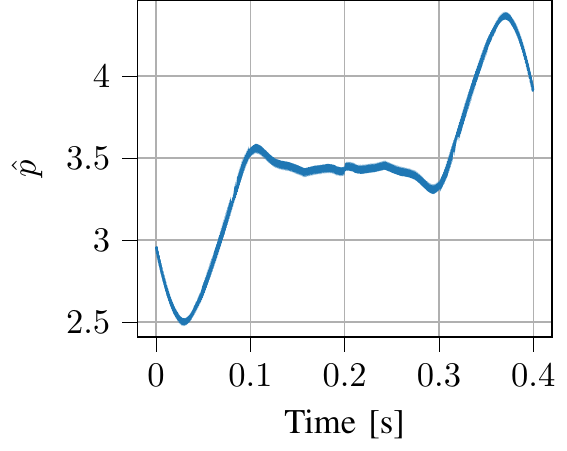}
	}
	\hspace{1.85cm}
	\subfloat[]{
		\includegraphics{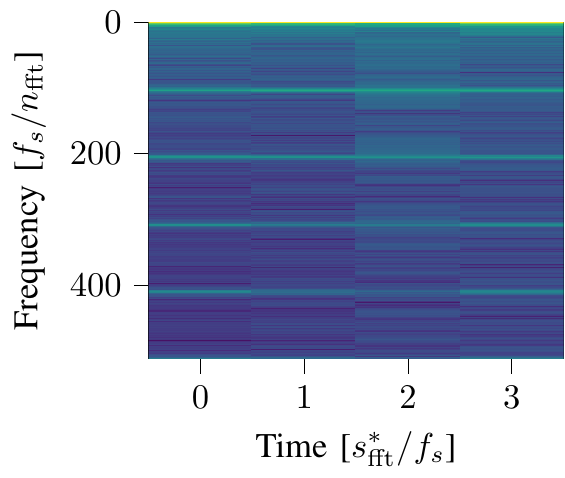}
	}
	\\
	\hspace{-1cm}
	\subfloat[]{
		\includegraphics{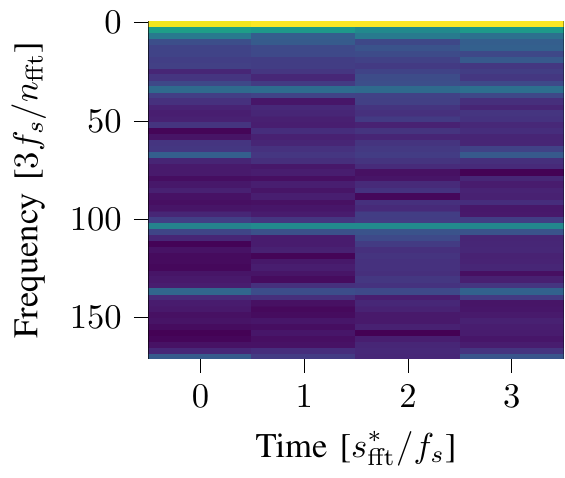}
	}
	\hspace{2cm}
	\subfloat[]{
		\includegraphics{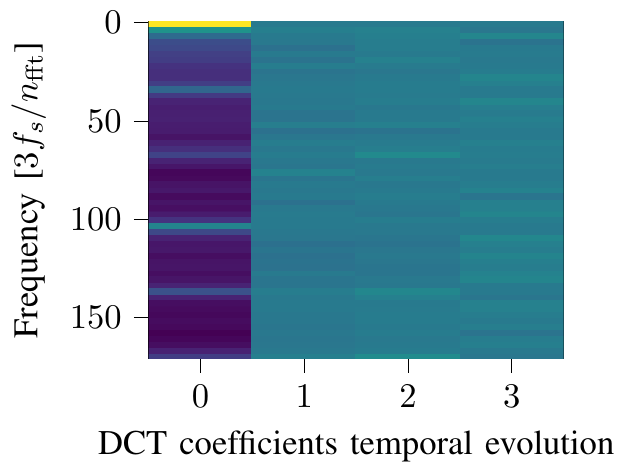}
	}
	\caption{Exemplary visualization of the 4 preprocessing steps. a) shows framed time data. The data is padded at its actual end ($t=\unit[0.2]{s}$). b) and c) show the original and compressed spectrograms, respectively. This visualization uses $s_{\text{fft}}^*=834$. The data in b) is additionally log-scaled to allow visualization. d) shows the final DCT coefficients (before flattening).}
	\label{fig:pipeline}
\end{figure*}

(2) When $n_c < n_f$, we pad the block with $n_f-n_c$ samples at its end by reflecting the samples at the edge value both in the direction of the time and amplitude axis, see \cref{fig:pipeline} a). This way, the padded frame is continuous at the edge points and resembles the original data in a repeated manner. Padding is only required during training to avoid mixing data with different failures when these failures occur in a very short time. Blocks of the validation data set are never padded. Instead, always the size of one frame is used, which means the block is extended to cover more noise data than originally specified by manual annotation.

\begin{figure}
	\centering
	\resizebox{7cm}{!}{\includegraphics{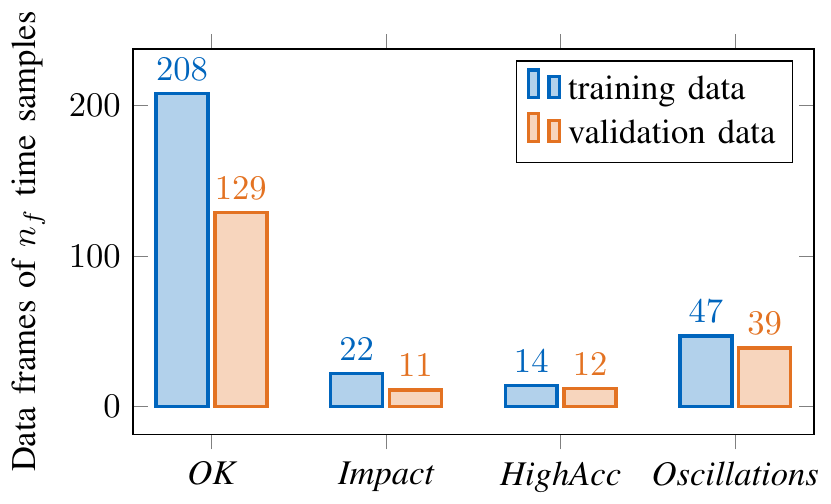}}
	\caption{The used data sets and their failure class distribution. Note that the events of \emph{Impact} and \emph{HighAcc} classes are usually very short, thus yielding a lower number of frames.}
	\label{fig:datashare}
\end{figure}

The output of this preprocessing step are $N_T=291$ training data frames of $n_f$ time samples each, and $N_V=191$ validation data frames. The failure class distribution of the data set is shown in \cref{fig:datashare}.

\subsubsection{Spectrogram Generation}

In the second step, we de-correlate the time series data by transformation to the frequency domain. This step helps to improve the performance of downstream machine learning approaches, which perform best for uncorrelated input data. To preserve information on the noise signal changes over time, a short-time Fourier transformation is performed on every frame. We use a $n_{\text{fft}}=1024$ points Fast Fourier transform (FFT) with Hann window, and a stride length $s_{\text{fft}}^*$, resulting in $n_t$ different transforms. The frame data is not padded at its end. We choose the FFT length based on the minimum resolvable frequency $f_{min} = \frac{f_s}{n_{\text{fft}}} \approx \unit[9.8]{Hz}$ --- a value where typically the lowest mechanical eigenfrequencies of robots reside, \cite{Berninger2020ICRA}. The amplitudes resulting from the FFTs are used for the spectrogram; phase data is not used. The output of this preprocessing step is a [$N_T \times n_{t} \times 513$] training data tensor / [$N_V \times n_{t} \times 513$] validation data tensor.

\subsubsection{Frequency Domain Compression}

The resolution of the fourier transform in frequency domain is $\frac{f_s}{n_{\text{fft}}} \approx \unit[9.8]{Hz}$, which leads to $\frac{n_{\text{fft}}}{2}+1 = 513$ frequency bins for the frequency range $[0, \frac{f_s}{2}]$. Several tests of the whole classification architecture have shown this high resolution to be unnecessary for the classification task. Thus, we compress the frequency spectrum with a factor of $3$ by binning the frequency samples into $\frac{\frac{n_{\text{fft}}}{2}+1}{3} = 171$ linear-spaced bins. This step has shown to significantly improve the accuracy on the validation data set, as it assumably reduces overfitting the training data. The amplitudes of the spectrogram are log-scaled after compression. The output of this preprocessing step is a [$N_T \times n_{t} \times 171$] training data tensor / [$N_V \times n_{t} \times 171$] validation data tensor.

\subsubsection{Resolving Time-Locality}

The spectrogram contains information on the evolution of the frequency spectrum over time. When directly using this data as input for a classical machine learning approach, the input data for two identical events, which occur at different points in time, are different. Due to this (time-) locality of the data, finding a valid model for all possible timings is difficult. Therefore, we apply a 1D type-II Discrete Cosine Transform (DCT) in orthogonal form on every frequency row of the spectrogram:
\begin{align}
	\operatorname{dct} \colon \mathbb{R}^{n_t} &\to \mathbb{R}^{n_t}\\
	x_n & \mapsto X_n, \; \text{with} \notag \\
	X_0 &= \frac{1}{\sqrt{n_t}} \sum \limits^{n_t-1}_{n=0} x_n, \notag \\
	X_k &= \sqrt{\frac{2}{n_t}} \sum \limits^{n_t-1}_{n=0} x_n \cos \left [ \frac{\pi}{2} \left ( n + \frac{1}{2} \right ) k \right ], \notag \\
	k &= 1, ..., n_t-1. \notag
\end{align}
This way, time-locality is resolved and the temporal evolution is stored in form of DCT coefficients for every frequency bin. The final preprocessing step flattens the output to a [$N_T \times 171 \cdot n_t$] training data tensor / [$N_V \times 171 \cdot n_t$] validation data tensor.

\subsection{Support Vector Machine Classifier}

For the classification, we apply a classical machine learning approach, a Support Vector Machine (SVM). SVMs have the advantage to be effective in high dimensional input spaces, even when the number of features in input data is higher than the size of the dataset --- which is the case here. Furthermore, they are resource efficient and provide adequate training times for small datasets.

\subsubsection{Two-Class SVM Formulation}

We use the \emph{libsvm}\cite{libsvm} implementation of the C Support Vector Classification (C-SVC) approach\cite{Boser1992, Cortes1995}. Given a data set of $n$-dimensional vectors $\vx_i \in \mathbb{R}^n, i = 1,...,m$, and corresponding binary label data $\vy \in \mathbb{R}^m, y_i \in \{ 1, -1\}$, C-SVC looks for the weights $\vw \in \mathbb{R}^l$ and bias $b$ by solving the following primal optimization problem \cite{libsvm}:
\begin{align}
	\underset{\vw, b, \vxi}{\text{min}} \; \; \; &\frac{1}{2} \vw^T \vw + C \sum_{i=1}^m \xi_i \\
	\text{s.t.} \; \; \; &y_i(\vw^T \phi(\vx_i) + b) \ge 1 - \xi_i, \; \;	i = 1,...,m\\
	&\xi_i \ge 0, \; \; i = 1,...,m,
\end{align}
where $\phi(\vx_i) : \mathbb{R}^n \rightarrow \mathbb{R}^l$ maps into the higher-dimensional space $l > n$, and $C > 0$ is a regularization parameter with the constraint error variables $\xi_i$. The value of $C$ is inversely proportional to the $l_1$ regularization of the weights $\vw$, which helps to avoid overfitting the training data. Regularization of $\vw$ tries to keep the weights minimal, i.e. use a minimal set of weights to represent a decision boundary.

For a high dimensional $\vw$, it is computationally expensive or even unfeasible to solve the problem in its primal form. Instead, the corresponding dual formulation is used:
\begin{align}
	\underset{\valpha}{\text{max}} \; \; \; &-\frac{1}{2} \valpha^T \vQ \valpha + \sum_{i=1}^m \alpha_i \\
	\text{s.t.} \; \; \; &\vy^T \valpha = 0,\\
	&0 \le \alpha_i \le C, \; \;	i = 1,...,m,\\
	\text{where} \; \; \; &Q_{ij} = y_iy_jK(\vx_i, \vx_j) = y_iy_j \phi(\vx_i)^T \phi(\vx_j).
\end{align}
With this formulation, an explicit calculation of the high-dimensional vector $\phi(\vx_i)$ is no longer necessary. Instead, the scalar kernel function $K(\vx_i, \vx_j)$ is directly calculated from its arguments without explicit mapping to the $l$-dimensional space. At the optimal solution, furthermore, only a few of the lagrangian multipliers $\alpha_i$ are non-zero --- namely those, which correspond to the \emph{support vectors}. This set of input data vectors $\vx_i$ defines the decision boundaries in the input data hyperplanes.

Common choices for the kernel function are the linear kernel $K_{\text{linear}}(\vx_i, \vx_j) = \vx_i^T\vx_j$, and the radial basis function (RBF) kernel $K_{\text{rbf}} = exp \left ( - \gamma^* ||\vx_i - \vx_j||^2 \right )$.
From the solution of the dual problem and the primal-dual relationship, the decision function for classification of an input vector $\vx$ results:
\begin{align}
	\hat y = sgn \left ( \vw^T \phi(\vx) + b \right ) = sgn \left ( \sum_{i=1}^m y_i \alpha_i K(\vx_i, \vx) + b \right ).
\end{align}

\subsubsection{Multi-Class SVM Formulation}

To extend the binary SVM classification to multiple classes in the data, the ``one-vs-rest'' approach is deployed. For each class $cls$ in the dataset, a separate binary C-SVC is used to learn the appearance of the class in the data. The C-SVCs share the same parameters and kernel, except for the regularization parameters $C_{cls}$, which are set inversely proportional to the respective class frequency $\frac{N_T}{N_{T,cls}}$ of the training data set: $C_{cls} = \frac{1}{4} \frac{N_T}{N_{T,cls}}  \hat C^*$. The global regularization parameter $\hat C^*$ is subject to the hyperparameter search described in \cref{sec:noisytest:results}.

\section{Results}
\label{sec:noisytest:results}

\emph{NoisyTest} is implemented in \emph{python3}. It uses \emph{scikit-learn}'s\footnote{https://scikit-learn.org} wrapper to \emph{libsvm} \cite{libsvm} for the support vector machine, and \emph{TensorFlow}\footnote{https://tensorflow.org} for data preprocessing. The code \cite{noisytest} and used data set \cite{noisytest-data-lola} are available online.

\subsection{Hyperparameter Search}
\label{sec:hyperparametersearch}

\begin{figure}
	\centering
	\includegraphics{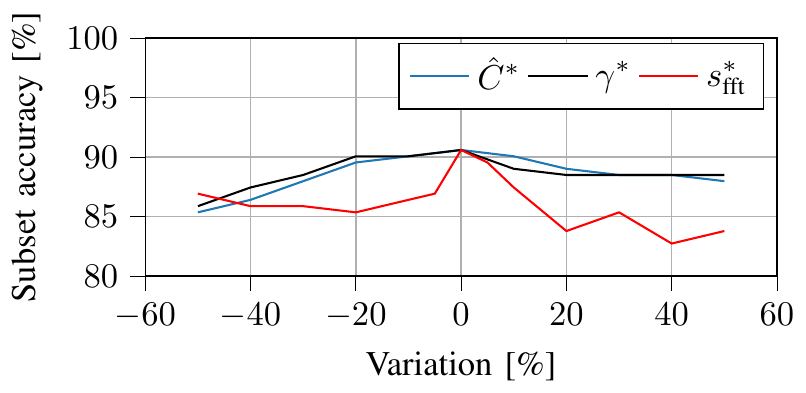}
	\caption{Accuracy for a variation of one parameter from its optimal value for the RBF kernel.}
	\label{fig:sensitivity}
\end{figure}

The preprocessing pipeline and the SVM implementation have a set of hyperparameters, which need to be tuned to the data set and application for maximum performance. These parameters are the FFT window stride length $s_{\text{fft}}^*$, the regularization parameter $\hat C^*$, and the kernel parameter $\gamma^*$ for radial basis function kernel. The best set of parameters maximizes the accuracy for all classes (subset accuracy) on the validation data set. It is found with a grid search by training the SVM for every parameter combination and evaluating the accuracy on the validation data set. For more details, refer to the implementation. The resulting optimal hyperparameters are shown in \cref{tab:hyperparameters}.
\begin{table}[ptb]
	\centering
	\begin{tabular}{@{}lll@{}}\toprule
		Parameter & Linear kernel & RBF kernel \\
		\midrule
		FFT stride length $s_{\text{fft}}^*$ & 401 & 834 \\
		Inv. regularization $\hat C^*$ & 1.0 & 1.1 \\
		Kernel parameter $\gamma^*$ & - & 5.7e-04 \\
		\midrule
		$n_t$ & 8 & 4 \\
		SVM input vector size & 1368 & 684 \\
		\bottomrule
	\end{tabular}
	\caption{Optimal hyperparameter values and terms resulting from this selection.}
	\label{tab:hyperparameters}
\end{table}
The sensitivity of the hyperparameters for the RBF kernel is shown in \cref{fig:sensitivity}.

\subsection{Accuracy and Discussion}

The accuracy of \emph{NoisyTest} is evaluated on the validation data set. Due to the relatively low sensivity of the hyperparameters (see \cref{fig:sensitivity}) and the small data set we are not using an extra test data set. The results for the linear kernel are summarized in \cref{tab:results:linear}; accuracies with the radial basis function kernel are shown in \cref{tab:results:rbf}.
\begin{table}[ptb]
	\centering
	\begin{tabular}{@{}lllll@{}}\toprule
		& \emph{OK} & \emph{Impact} & \emph{HighAcc} & \emph{Oscillations} \\
		\midrule
		False negative errors & \unit[0.0]{\%} & \unit[1.0]{\%} & \unit[5.8]{\%} & \unit[4.7]{\%}\\
		False positive errors & \unit[5.8]{\%} & \unit[5.8]{\%} & \unit[0.0]{\%} & \unit[0.0]{\%} \\
		\midrule
		Failure detection rate & \multicolumn{4}{c}{\unit[94.2]{\%}} \\
		Subset accuracy & \multicolumn{4}{c}{\unit[88.5]{\%}} \\
		\bottomrule
	\end{tabular}
	\caption{Achieved accuracies on the validation data set with the \textbf{linear} kernel}
	\label{tab:results:linear}
\end{table}
\begin{table}[ptb]
	\centering
	\begin{tabular}{@{}lllll@{}}\toprule
		& \emph{OK} & \emph{Impact} & \emph{HighAcc} & \emph{Oscillations} \\
		\midrule
		False negative errors & \unit[0.5]{\%} & \unit[2.6]{\%} & \unit[6.3]{\%} & \unit[0.0]{\%}\\
		False positive errors & \unit[2.1]{\%} & \unit[2.6]{\%} & \unit[0.0]{\%} & \unit[4.7]{\%} \\
		\midrule
		Failure detection rate & \multicolumn{4}{c}{\unit[97.9]{\%}} \\
		Subset accuracy & \multicolumn{4}{c}{\unit[90.6]{\%}} \\
		\bottomrule
	\end{tabular}
	\caption{Achieved accuracies on the validation data set with the \textbf{RBF} kernel}
	\label{tab:results:rbf}
\end{table}
Both approaches perform well in terms of the overall detection of failures (\emph{OK} vs. rest). The RBF kernel performs best in terms of subset accuracy. However, it shows a non-zero false-negative rate for the \emph{OK} class, i.e., a non-zero false-negative rate for the detection of a failure. We focus on the results of the RBF kernel variant in the following. From the shown accuracy data --- and additional evaluations with reduced validation data set --- we found that the undetected failures are caused by the \emph{Impact} class of scenario 21. Furthermore, the \emph{HighAcc} events in scenario 24 are falsely detected as \emph{Impact} and \emph{Oscillations} classes. The torque signal jumps induced in this scenario, however, are hard to classify as closely related to the effects of an impact and as they certainly also lead to subsequent oscillations. In reality, it is somehow likely that the classes sometimes only occur together or in close succession. Nevertheless, identifying the possible failure class (instead of just a single-class evaluation) may provide the user with additional hints on the failure cause.
We conducted a test to evaluate if the SVM learns the concrete features of a failure independently of robot-specific noise characteristics. For this check, we generated a reduced noise estimate for \lola based not on all joints, but only on the knee and ankle joints of both feet. The reduced number of involved joints results in a completely different time-domain signal. Evaluating the trained SVM with RBF kernel on a separated validation set with the reduced noise estimate still gives reasonable detection performance, see \cref{tab:results:rbfreduced}. This result indicates the possibility of using a single trained model for different types of robots.
\begin{table}[ptb]
	\centering
	\begin{tabular}{@{}lllll@{}}\toprule
		& \emph{OK} & \emph{Impact} & \emph{HighAcc} & \emph{Oscillations} \\
		\midrule
		False negative errors & \unit[1.6]{\%} & \unit[4.2]{\%} & \unit[6.3]{\%} & \unit[7.3]{\%}\\
		False positive errors & \unit[9.4]{\%} & \unit[7.9]{\%} & \unit[0.0]{\%} & \unit[2.1]{\%} \\
		\midrule
		Failure detection rate & \multicolumn{4}{c}{\unit[90.6]{\%}} \\
		Subset accuracy & \multicolumn{4}{c}{\unit[80.6]{\%}} \\
		\bottomrule
	\end{tabular}
	\caption{Achieved RBF-kernel SVM accuracies on the validation data set with reduced noise estimate. Only the ankle and knee joints of both feet are used.}
	\label{tab:results:rbfreduced}
\end{table}
\section{Conclusion}
\label{sec:noisytest:conclusion}
This paper presents the idea and implementation of \emph{NoisyTest}, a tool for virtual system tests based on the estimated noise of a robot. We show that machine learning techniques on a scalar noise signal, generated from the simulated robot movement, detect different failures with high accuracy. The results show the generalizability of the approach to different robot types, even when the same trained model is used. \emph{NoisyTest} is primarily intended for simulated system tests to help speed up the development of robot control software by finding failures in simulation data, which are hard to identify by manual inspection and are often only recognized during real-world experiments. It helps to generate automated system test pipelines with a more precise definition of success than conventional heuristic methods can offer.

Currently, we use this technique for a general test of simulation runs. However, several extensions are considered for the future. First, it might be possible to run \emph{NoisyTest} on several noise estimates generated from intermediate signals in the control algorithm's data flow to localize the faulty code. Second, it is interesting to test its performance on noise estimates generated from experimental data. A possible use case is a fully automatic system test of the actual robot system with a noise-based evaluation of its performance. Furthermore, the used classification technique is computationally efficient, presumably allowing online real-time diagnostics of the robot.

We strive for the extension of the failure classes and the application to more robots. We invite other researchers to use \emph{NoisyTest} on their system and contribute to a shared tool for failure detection in robotic system development.

\enlargethispage{-4.3cm}    

\bibliographystyle{IEEEtran}
\bibliography{references}
\end{document}